\documentclass[10pt,twocolumn,letterpaper]{article}

\usepackage{iccv}
\iccvfinalcopy

\usepackage{times}

\usepackage{epsfig}
\usepackage{graphicx}
\usepackage{amsmath}
\usepackage{amssymb}
\usepackage{gensymb}

\usepackage[utf8]{inputenc} 
\usepackage[T1]{fontenc}    
\usepackage{url}            
\usepackage{booktabs}       
\usepackage{amsfonts}       
\usepackage{nicefrac}       
\usepackage{microtype}      

\usepackage{subcaption}
\usepackage{changepage}
\usepackage{epstopdf}
\usepackage{xcolor}
\usepackage{colortbl}
\usepackage{mathtools}
\usepackage{amsmath}
\usepackage{setspace}
\usepackage{booktabs}
\usepackage{array}
\usepackage{algorithm}
\usepackage{algorithmic}
\usepackage{multirow}
\usepackage{wrapfig}
\usepackage{bbm}
\usepackage{blindtext}
\usepackage{arydshln}

\usepackage{float}

\usepackage{pifont}
\usepackage[symbol]{footmisc}

\usepackage[pagebackref=true,breaklinks=true,colorlinks,bookmarks=false]{hyperref}

\def\iccvPaperID{8116} 

\usepackage{amsmath}
\usepackage{amssymb}
\usepackage{amsfonts}
\usepackage{mathtools}
\usepackage{bm}
\usepackage{xfrac}

\definecolor{darkgreen}{rgb}{0.0, 0.5, 0.13}

\newcommand{\chrisnote}[1]{{  \color{red} [[ #1 -- Chris ]] }}

\newcommand{\matt}[1]{\textcolor{orange}{[\textbf{Matt:}] #1}}

\newcommand{\keunhong}[1]{\textcolor{blue}{[\textbf{Keunhong:} #1}}

\definecolor{myblue}{rgb}{0, 0.531, 0.9}

\newcommand{\bb}[1]{\mathbf{#1}}

\newcommand{\expec}{\mathbb{E}}

\def\bal#1\eal{\begin{align*}#1\end{align*}}

\newcommand{\removed}[1]{}

\newcommand{\cmark}{\ding{51}}
\newcommand{\xmark}{\ding{55}}

\newcommand{\mat}[1]{\bm{\mathrm{#1}}}

\newcommand{\imagefidelitymetric}[1]{\textcolor{orange}{#1}}
\newcommand{\heldoutmetric}[1]{\textcolor{cyan}{#1}}

\newcommand{\website}{\url{fig-nerf.github.io}}

\definecolor{myyellow}{rgb}{1,1, 0.6}
\definecolor{myorange}{rgb}{1, 0.8, 0.6}
\definecolor{myred}{rgb}{1, 0.6, 0.6}
\newcommand{\tablefirst}[0]{\cellcolor{myred}}
\newcommand{\tablesecond}[0]{\cellcolor{myyellow}}
\newcommand{\tablethird}[0]{}

\newcommand{\nerfl}{NeRF+L}
\newcommand{\nerfls}{NeRF+L+S}
\newcommand{\nerflsd}{NeRF+L+S+D}

\newcommand{\postfigure}{\vspace{0mm}}  
\newcommand{\postsubsubsec}{\vspace{0mm}}  
\newcommand{\postsubsec}{\vspace{0mm}}  

\begin{document}


\title{FiG-NeRF: Figure-Ground Neural Radiance Fields for 3D Object Category Modelling}


\author{
Christopher Xie\textsuperscript{1 $\dagger$} \and Keunhong Park\textsuperscript{1 $\dagger$} \and Ricardo Martin-Brualla\textsuperscript{2} \and Matthew Brown\textsuperscript{2} \smallskip \and
\textsuperscript{1}University of Washington\qquad 
\textsuperscript{2}Google Research \\[6pt]
\website
}

\maketitle

\footnotetext[2]{Work done while the author was an intern at Google.}

\begin{abstract}
We investigate the use of Neural Radiance Fields (NeRF) to learn high quality 3D object category models from collections of input images.
In contrast to previous work, we are able to do this whilst simultaneously separating foreground objects from their varying backgrounds. 
We achieve this via a 2-component NeRF model, FiG-NeRF, that prefers explanation of the scene as a geometrically constant background and a deformable foreground that represents the object category.
We show that this method can learn accurate 3D object category models using only photometric supervision and casually captured images of the objects.
Additionally, our 2-part decomposition allows the model to perform accurate and crisp amodal segmentation.
We quantitatively evaluate our method with view synthesis and image fidelity metrics, 
using synthetic, lab-captured, and in-the-wild data. Our results demonstrate convincing 3D object category modelling that exceed the performance of existing methods.
\end{abstract}



\section{Introduction}

Learning high quality 3D object category models from visual data has a variety of applications such as in content creation and robotics. For example, convincing category models might allow us to generate realistic new object instances for graphics applications, or allow a robot to understand the 3-dimensional structure of a novel object instance if it had seen objects of a similar type before~\cite{agnew2020amodal,Wang_2019_CVPR}. Reasoning about objects in 3D could also enable improved performance in general perception tasks. For example, most work in object detection and instance segmentation is limited to learning object categories in 2D. Using 3D category models for such tasks could enable enhanced reasoning, such as amodal segmentation~\cite{zhu2017semantic} 
taking into account occlusions of multiple objects, or fusion of information over multiple views taken from different viewpoints 

\begin{figure}[t]
    \centering
    \includegraphics[width=\linewidth]{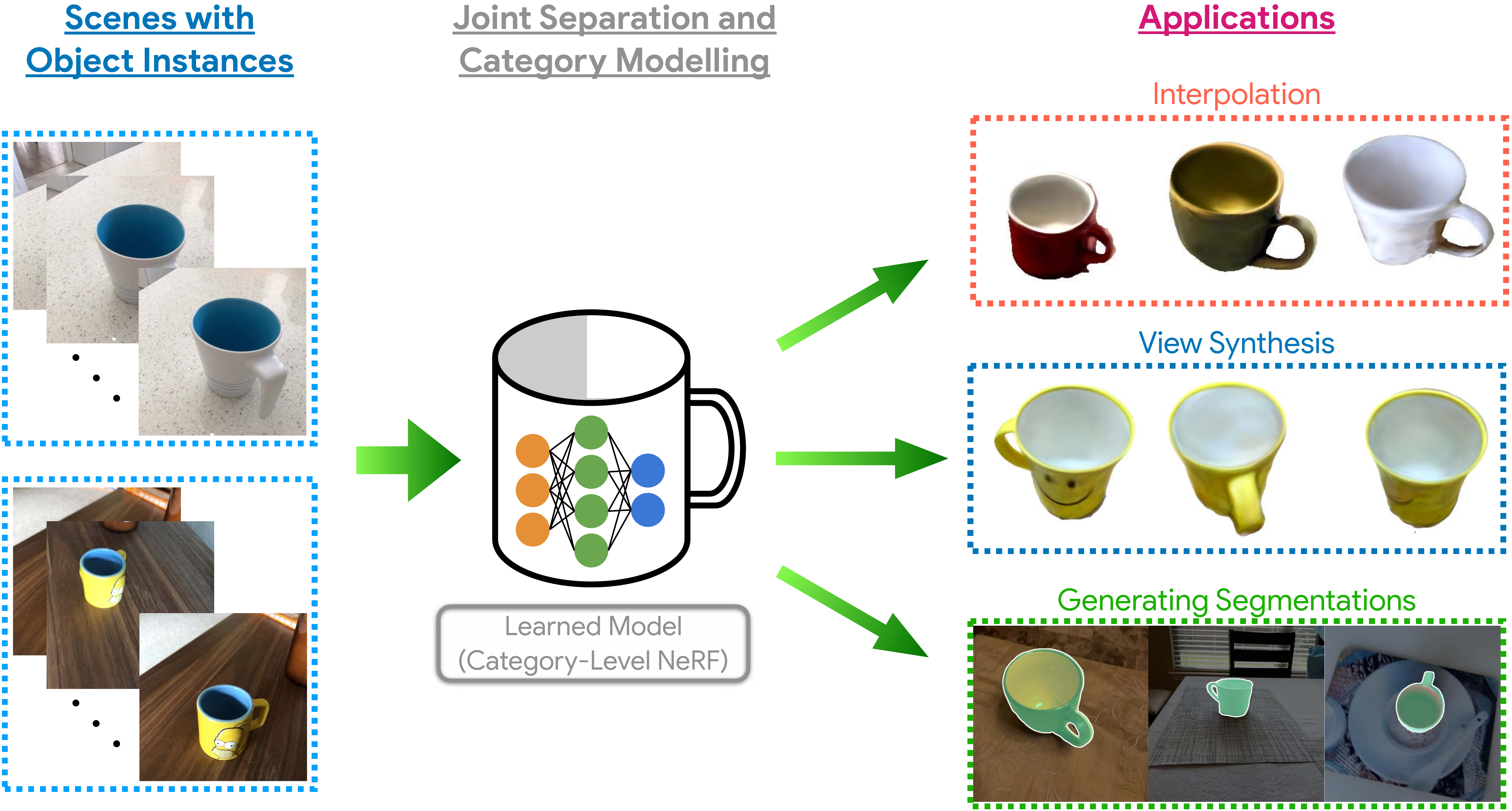}
    \caption{
    Overview of our system. We take as input a collection of RGB captures of scenes with objects of a category. Our method jointly learns to decompose the scenes into foreground and background (without supervision) and a 3D object category model, that enables applications such as instance interpolation, view synthesis, and segmentation.
    }
    \label{fig:teaser}
    \postfigure
\end{figure}

The majority of existing work in 3D category modelling from images has used supervision in the form of 3D models~\cite{shapenet2015}, segmentations~\cite{WilesZ17,henzler2021unsupervisedvideos} or semantic keypoints~\cite{cmrKanazawa18}. 
Recently, some authors have attempted to learn 3D category models using images alone~\cite{HoloGAN2019,schwarz2020graf}; however, these methods assume simple backgrounds or known silhouettes.
An open problem is to learn 3D object category models using casually captured photography with unconstrained backgrounds and minimal supervision.
In this work, we pursue this objective, using the recently released Objectron dataset~\cite{objectron2020} as a target for in-the-wild 3D object category modelling.
In this setting, backgrounds are different for each instance, so our method must separate the object from its background as well as understand its particular shape and appearance.
 
We build on top of Neural Radiance Fields (NeRF)~\cite{mildenhall2020nerf}, which has shown excellent results for image-based view synthesis. We propose Figure-Ground Neural Radiance Fields (FiG-NeRF),
which uses two NeRF models to model the objects and background, respectively.
To enable separation of object (figure) from background (ground, as in the Gestalt principle of figure-ground perception), we adopt a 2-component model comprised of a deformable foreground model~\cite{park2020nerfies} and background model with a fixed geometry and variable appearance.
We find that fixing the geometry of the background is often appropriate for the object categories we study. For example, cups typically rest on tables, and eyeglasses on faces.
We show that our 2-component approach together with sparsity priors is sufficient for the model to successfully separate modelling of a foreground object and background.
Since our model infers this separation between object and background, a secondary benefit, in addition to the 3D object category model, is a crisp amodal object/background segmentation. In our evaluation we show that the quality of our segmentations outperform both Mask R-CNN~\cite{he2017mask} and a bespoke matting algorithm~\cite{martinbrualla2020gelato} that uses additional images.

The key contributions of our work are: 
\begin{enumerate}
    \item We jointly estimate object category models whilst separating objects from their backgrounds, using a novel 2-component, Deformable NeRF formulation. 
    \item Our results in novel-view synthesis and instance interpolation outperforms baselines such as non-deformable NeRF variants and SRNs~\cite{sitzmann2019srns} on both synthetic and real image datasets.
    \item We demonstrate learning of object category models in-the-wild and with variable backgrounds, using the Objectron dataset~\cite{objectron2020}.
\end{enumerate}

To our knowledge, our method is the first to jointly estimate
detailed 
object category models with backgrounds of variable and view-dependent appearance, without recourse to silhouettes or 2D segmentation predictions.

\section{Related Work}


Early work in 3D object category modelling reconstructed objects from the PASCAL VOC dataset~\cite{vicente2014reconstructing}.
Rigid SFM was used to find a mean shape, followed by intersection of a carefully selected subset of silhouettes. Kar et al.~\cite{kar2015category} also uses silhouette supervision, employing non-rigid SFM, followed by fitting a shape with mean and deformation modes.


More recent investigations adapt classical 3D modelling representations into a deep learning framework.
Pixel2Mesh~\cite{wang2018pixel2mesh} uses an iterative mesh refinement strategy to predict 
a 3D triangle mesh from the input image.
The method uses ground truth mesh supervision, deforming an initial 156 vertex ellipsoid via a series of graph convolutions and coarse-to-fine mesh refinement. 
Category Specific Mesh reconstruction~\cite{cmrKanazawa18} also works by deformation of a fixed initial mesh, but without requiring 3D ground truth. It uses silhouette and keypoint based loss terms,
with a differentiable mesh renderer~\cite{kato2018renderer} employed to calculate gradients.


The above mesh methods are limited a spherical topology. Mesh-RCNN~\cite{gkioxari2019mesh} removes this limitation and provides more flexible shape modelling by initially predicting a voxel grid in a manner similar to ~\cite{choy20163d}, before adapting this towards a mesh model.
The voxel-based 3D reconstruction has limited ability to model fine detail due to the high memory cost in storing a large voxel array. This is partly addressed in ~\cite{ogn2017}, which use octrees to enable resolutions up to $512^3$. 



More recent work fuses classical computer graphics techniques with deep generative models~\cite{sitzmann2019deepvoxels,tewari2020state,thies2019deferred,park2019latentfusion}. A promising recent trend involves the use of coordinate regression approaches, also known as Compositional Pattern Producing Networks (CPPNs)~\cite{stanley2007compositional}. These are fully connected networks operating on 3D coordinates to produce an output that represents the scene. Examples include DeepSDF~\cite{park2019deepsdf}, which predicts a signed distance function at each 3D coordinate. 
Scene Representation Networks~\cite{sitzmann2019srns} (SRNs) represent an embedding vector at each 3D position, with a learned recurrent renderer mimicking the role of sphere tracing.
NeRF~\cite{mildenhall2020nerf} also uses CPPNs, but within a volumetric rendering scheme. 
A novel positional encoding is key to the success of this approach~\cite{sitzmann2019siren,tancik2020fourfeat}.


While much prior work makes use of supervision in the form of 3D ground truth, keypoints or silhouettes,~\cite{novotny2017learning} learns 3D shape using only image inputs, estimating viewpoint and depth, but limited to a point cloud reconstruction inferred from the depth map.~\cite{wu2020unsupervised} uses a symmetry prior and factored representation to generate improved results.
Generative adversarial nets have also been used to model objects from image collections alone, using 3D convolutions~\cite{gadelha20173d,henzler2019escaping}, and 3D+2D convolutions in~\cite{HoloGAN2019}, the latter giving more realistic results at the expense of true 3D view consistency. Shelf-supervised mesh prediction~\cite{ye2021shelf} combines image segmentation inputs with an adversarial loss, generating impressive results over a broad range of image categories.


Close to our approach, Generative Radiance Fields~\cite{schwarz2020graf} use a latent-variable conditioned NeRF to model object categories, trained via adversarial loss. They are however, limited to plain backgrounds. GIRAFFE~\cite{niemeyer2020giraffe} extends this with a composition of NeRF-like models that generate a ``feature field”, rendered via a 2D neural render to enable controllable image synthesis. They also show results in separating object from background, though since the feature images are only 16x16 (upsampled via a 2D neural render), the outputs are blurry and lack the fine geometric detail produced by our method. 
STaR~\cite{yuan2021star} also performs foreground and background separation with NeRFs, but only considers a single rigidly-moving object.
\section{Method}
\label{sec:methods}

In this section, we introduce our model, Figure-Ground NeRF (FiG-NeRF), which is named after the Gestalt principle of figure-ground separation~\cite{ren2006figure}.

\begin{figure}[t]
    \centering
    \includegraphics[width=\linewidth]{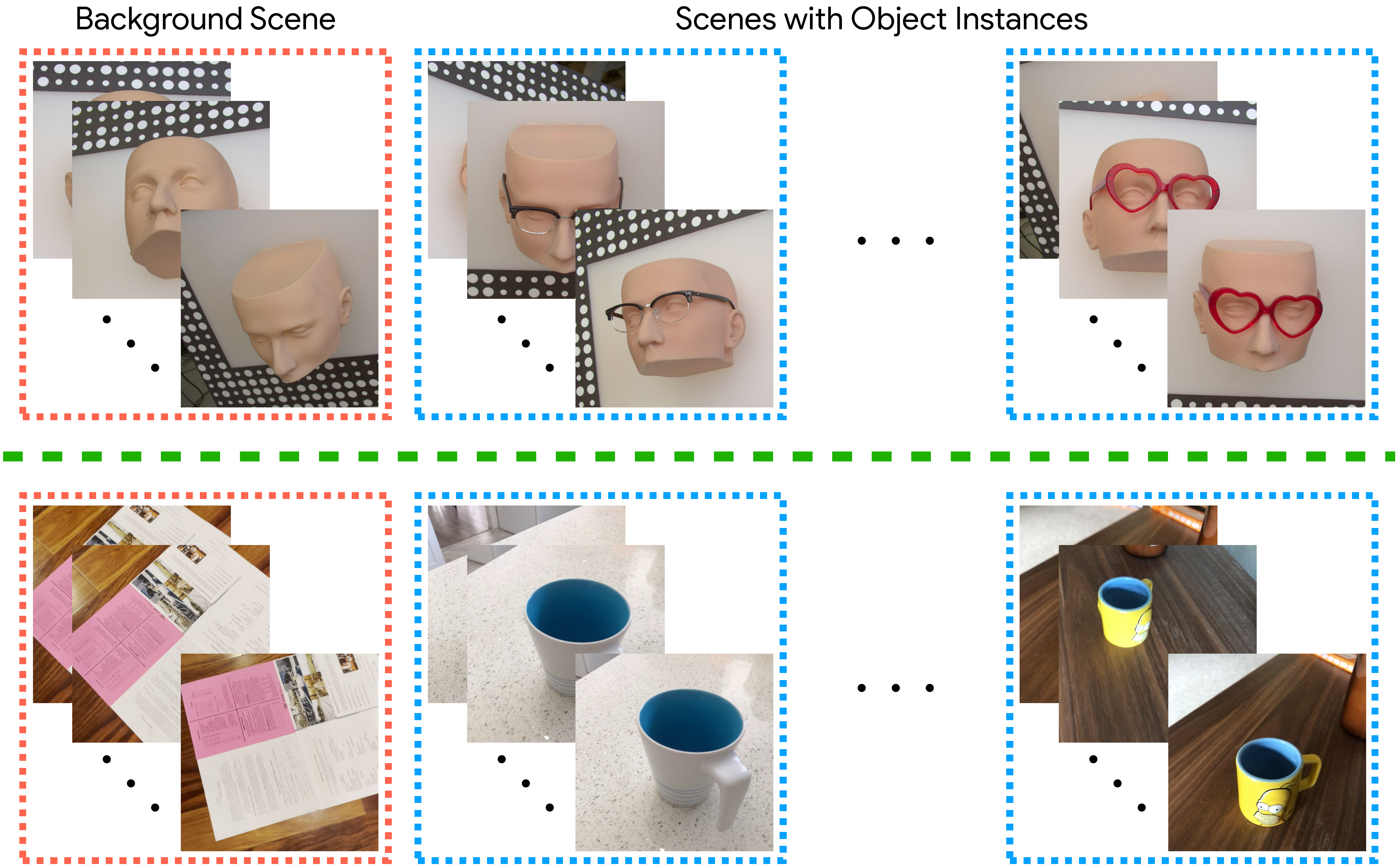}
    \caption{Example setups for \textsc{\small Glasses} (top) and \textsc{\small Cups} (bottom) datasets. For the lab-captured \textsc{\small Glasses} dataset~\cite{martinbrualla2020gelato}, the \textcolor{red}{background} (left) is a mannequin, and each \textcolor{myblue}{scene} (right) is a different pair of glasses placed on the mannequin. For \textsc{\small Cups}, we build this from the Objectron~\cite{objectron2020} dataset of crowdsourced casual cellphone video captures, where we captured a planar surface with textures (colored papers) for the background.}
    \label{fig:dataset_setup}
    \postfigure
\end{figure}

\subsection{Setup}

We consider a setting with a collection of $N$ scenes, where each scene contains a single object instance. We assume that the scene backgrounds have identical geometry but not necessarily color. For example, different tabletops/planar surfaces, or the same scene with different lighting satisfy this assumption. Each instance $i$ is captured from $M_i$ viewpoints, resulting in a dataset $D_i = \{I_i^j\}_{j=1}^{M_i}$ for each instance $i = 1, \ldots, N$. Additionally, we have $M_\text{bg}$ images of a background scene with no object instance present, and designate this as $D_\text{bg}$. We denote the entire dataset as $\mathcal{D} = \left\{\cup_{i=1}^N D_i \right\} \cup  D_\text{bg}$. We also assume that we have camera poses and intrinsics for each image. This can be obtained by standard structure-from-motion (SfM) software such as \cite{schoenberger2016sfm} or by running visual odometry during capture~\cite{objectron2020}. See Figure \ref{fig:dataset_setup} for an example of our assumed setup.

\subsection{Preliminaries}

A Neural Radiance Field (NeRF) \cite{mildenhall2020nerf} is a function $F : (\bb{x}, \bb{d}) \rightarrow (\bb{c}, \sigma)$ comprised of multi-layer perceptrons (MLP) that map a 3D position $\bb{x}$ and 2D viewing direction $\bb{d}$ to an RGB color $\bb{c}$ and volume density $\sigma$. $F$ is comprised of a trunk, followed by separate color and density branches. Point coordinates are passed through a positional encoding layer $\gamma(\cdot)$ before being input to the MLP. The radiance field is rendered using volume rendering to generate highly realistic results when trained on many views.

A drawback to NeRF is that it only learns to model a single scene. To extend NeRF to handle multiple scenes, previous works have introduced conditioning input latent codes to model object categories~\cite{schwarz2020graf} and the different appearances of a scene~\cite{martinbrualla2020nerfw}. They concatenate the inputs of the NeRF MLPs with additional latent codes that are sampled using a GAN~\cite{goodfellow2014gan} or optimized during training~\cite{bojanowski2017optimizing}. We dub such models \textit{conditional NeRFs}. Our approach builds upon conditional NeRF models to handle object categories.




\subsection{Object and Background Separation}
\label{sec:decomposition}

\begin{figure}[t]
    \centering
    \includegraphics[width=\linewidth]{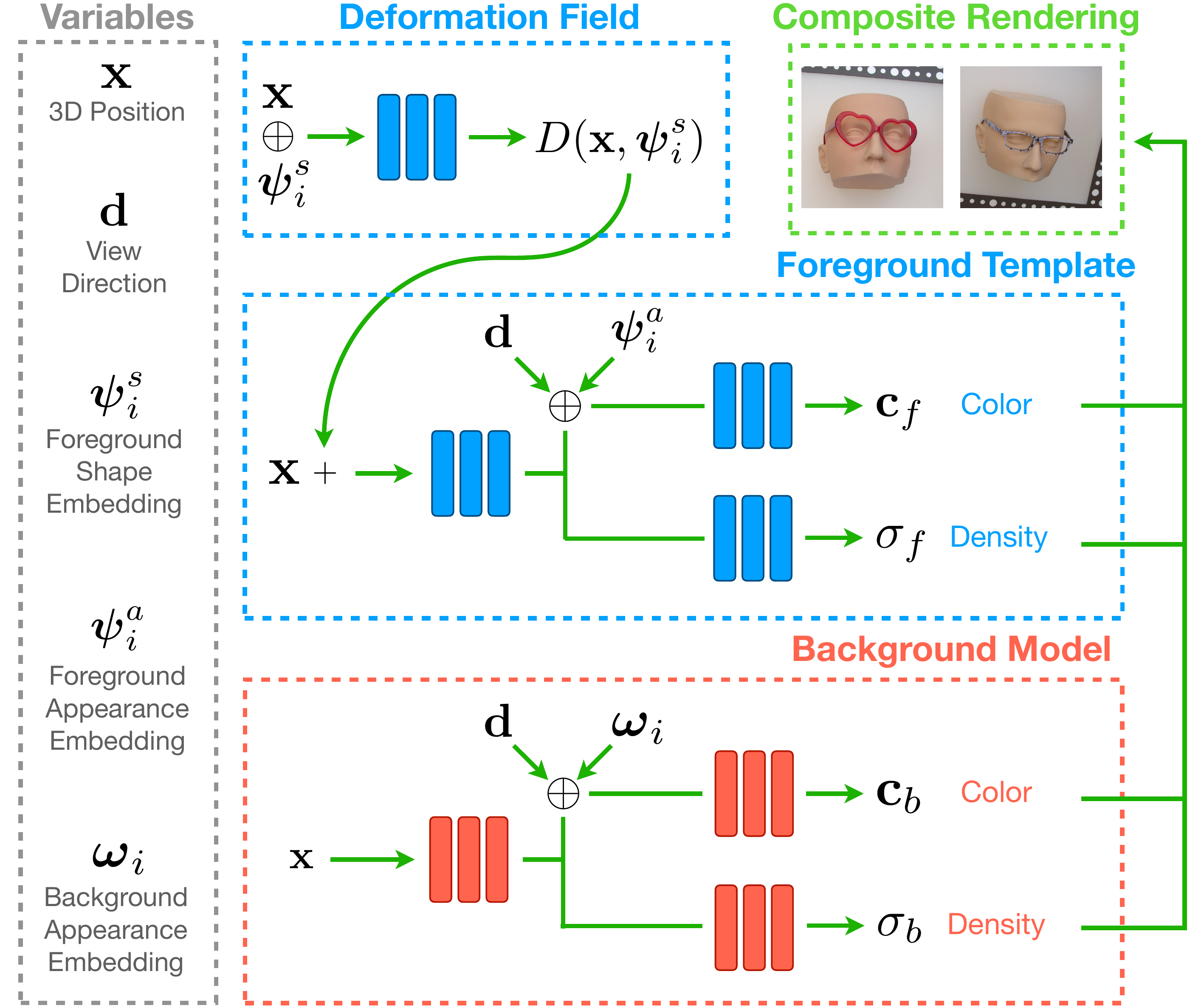}
    \caption{FiG-NeRF consists of \textcolor{myblue}{foreground} and \textcolor{red}{background} models. The \textcolor{myblue}{foreground} model, which includes the deformation field and template NeRF, is shown in blue while the \textcolor{red}{background} NeRF is shown in red. $\oplus$ denotes concatenation. }
    \label{fig:model_architecture}
    \postfigure
\end{figure}

Many previous works on category modelling rely on ground truth segmentations that define the extent of the objects. These can be easily extracted for synthetic datasets, but it is difficult to obtain accurate segmentations for real datasets. Additionally, this limits the potential categories to classes that the object detector has trained on.
Instead, we propose a model that learns a segmentation of the scene into a foreground component containing the object of interest, and a background component in an unsupervised manner.

The key to our approach is to decompose the neural radiance field into two components: a foreground component $F^f$ and a background component $F^b$, each modeled by a conditional NeRF. The foreground latent code $\boldsymbol{\psi}_i$ and the background latent code $\boldsymbol{\omega}_i$ condition the foreground and background component, respectively.

We observe that many objects are often captured in backgrounds containing approximately the same geometry, such as cups on planar surfaces, or eyeglasses on faces.
Our formulation exploits this observation, and assumes that all the object instances in $\mathcal{D}$ are captured against backgrounds that share the same geometry, while their appearance might change due to texture or lighting changes. 
Such an assumption is not overly restrictive, and allows for distributed data capture, i.e. each scene can be captured by a different user (for example, tabletops in different homes), as is the case of the Objectron dataset~\cite{objectron2020}.
We incorporate this inductive bias into the network by 
feeding $\boldsymbol{\omega}_i$ to the branch of $F^b$ that affects the appearance only, not density (geometry).
Note that with this formulation, the background model can capture background surface appearance variations induced by the object, such as shadows. 

\begin{table*}[h!]
\centering
\resizebox{\linewidth}{!}{\begin{tabular}{c||c:cccc|c:cccc|c:cccc}
\toprule
\multirow{2}{*}{Model} & \multicolumn{5}{c|}{\textsc{\small Cars}} & \multicolumn{5}{c|}{\textsc{\small Glasses}} & \multicolumn{5}{c}{\textsc{\small Cups}} \\
\cline{2-16} & \imagefidelitymetric{FID} $\downarrow$ & \heldoutmetric{PSNR}$\uparrow$ & \heldoutmetric{SSIM}$\uparrow$ & \heldoutmetric{LPIPS}$\downarrow$ & \heldoutmetric{IoU}$\uparrow$ & \imagefidelitymetric{FID}$\downarrow$ & \heldoutmetric{PSNR}$\uparrow$ & \heldoutmetric{SSIM}$\uparrow$ & \heldoutmetric{LPIPS}$\downarrow$ & \heldoutmetric{IoU}$\uparrow$ &\imagefidelitymetric{FID}$\downarrow$ & \heldoutmetric{PSNR}$\uparrow$ & \heldoutmetric{SSIM}$\uparrow$ & \heldoutmetric{LPIPS}$\downarrow$ & \heldoutmetric{IoU}$\uparrow$ \\ \hline
SRNs~\cite{sitzmann2019srns} & 34.96 & 36.42 & .9860 & .0142 & -- & -- & -- & -- & -- & -- & 213.8 & 17.20 & 0.6168 & 0.6580 & -- \\
\nerfl{} & \tablesecond 34.86 & \tablethird 37.75 & \tablesecond .9884 & \tablethird .0112 & -- & \tablethird 43.37 & \tablethird 36.17 & \tablethird .9390 & \tablethird .1020 & -- & \tablethird 164.0 & \tablethird 24.40 & \tablethird .9402 & \tablethird .0758 & -- \\
\nerfls{} & \tablethird 35.48 & \tablesecond37.78 & \tablethird.9882 & \tablesecond.0107 & \tablesecond.9555 & \tablefirst39.17 & \tablesecond36.24 & \tablesecond.9397 & \tablesecond.0986 & \tablesecond.5572 & \tablesecond126.0 & \tablefirst 25.10 & \tablefirst.9437 & \tablesecond.0666 & \tablesecond.8023 \\
\nerflsd{} & \tablefirst26.02 & \tablefirst38.02 & \tablefirst.9889 & \tablefirst.0097 & \tablefirst.9590 & \tablesecond39.56 & \tablefirst36.26 & \tablefirst.9402 & \tablefirst.0968 & \tablefirst.5796 & \tablefirst 106.4 & \tablesecond 25.05 & \tablesecond.9430 & \tablefirst.0651 & \tablefirst.8535 \\ 
\bottomrule
\end{tabular}
}
\caption{Quantitative results on all datasets. We show \imagefidelitymetric{instance interpolation} metrics and \heldoutmetric{heldout view} metrics. Our proposed method outperforms the baselines on virtually every metric. We color code each row as \colorbox{myred}{{best}} and \colorbox{myyellow}{second best}.}
\label{table:model_comparison}
\end{table*}

More formally, our model is composed of the conditional NeRF models:
\begin{subequations}
\label{eq:base_model}
\begin{align}
F^f &: \left(\bb{x}, \bb{d}, \boldsymbol{\psi}_i \right) \rightarrow \left(\bb{c}_f, \sigma_f\right)\\
F^b &: \left(\bb{x}, \bb{d}, \boldsymbol{\omega}_i\right) \rightarrow \left(\bb{c}_b, \sigma_b\right)\,,
\end{align}
\end{subequations}
where we learn the foreground and background latent codes using generative latent optimization (GLO)~\cite{bojanowski2017optimizing}.

To render images using the 2-component model, we follow the composite volumetric rendering scheme of \cite{martinbrualla2020nerfw}. That is, for a ray $\bb{r}(t) = \bb{x} + t\bb{d}$, we compute the color as
\begin{equation}
    \bb{C}(\bb{r}) = \int_{t_n}^{t_f} T(t) \left( \sigma_f(t) \bb{c}_f(t) + \sigma_b(t) \bb{c}_b(t) \right) dt \,,
\end{equation}
where $T(t) = \exp\left( - \int_{t_n}^t \left(\sigma_f(s) + \sigma_b(s) \right) ds \right)$, and $t_n, t_f$ are the near and far integration bounds, respectively. $\sigma_k(t), \bb{c}_k(t)$ are the density and color values at $\bb{r}(t)$ for $k \in \{f,b\}$, respectively.

With this separation, we can compute a foreground segmentation mask by rendering the depth of the foreground and background models separately and selecting the pixels in which the foreground is closer to the camera than background. We show in Section~\ref{subsubsec:segmentation_results} that our learned model produces accurate and crisp amodal segmentation masks.

\subsection{Objects as Deformed Template NeRFs}

We would like our category models to (1) allow for smooth interpolation between objects, and (2) be robust to partial observations of specific instances. Understanding these goals, we make the observation that instances of an object category are typically structurally very similar. For example, it is possible to think of a generic cup, shoe, or camera without specifying a specific instance. This observation has been used to motivate methods such as \emph{morphable models}~\cite{blanz1999morphable} which deform a canonical mesh to fit objects with small intra-class variations e.g., faces.  Inspired by the success of such methods, we propose to model object instances as deformations from a canonical radiance field. While morphable models assume a shared topology and vertex structure, we are not bound by these limitations due to our use of a continuous, volumetric representation i.e., NeRF.

We incorporate canonical object templates into our model by adapting the concept of Deformable NeRF or \emph{nerfies}~\cite{park2020nerfies} to modelling object categories. Deformable NeRFs are comprised of two parts: a canonical template NeRF $G$, which is a standard 5D NeRF model, and a deformation field $D$ which warps a point $\bb{x}$ in observation-space coordinates to a point $\bb{x}'$ in template-space. The deformation field $(\bb{x}, \boldsymbol{\delta}_t) \mapsto \bb{x}'$ is a function conditioned by a deformation code $\boldsymbol{\delta_t}$ defined for time step $t$.
We represent it with a residual translation field $D$ such that $\bb{x}' = \bb{x} + D(\mat{x}, \boldsymbol{\delta}_t)$ where $D$ is a coordinate-based MLP that uses a positional encoding layer.

Instead of associating deformation fields to time steps $t$, our model associates a deformation field to each object instance $i$ represented by a shape deformation code $\boldsymbol{\psi}_i^s$. Because all objects share the template NeRF model $G$ yet may have different textures, we condition $G$ with a per-instance appearance code $\boldsymbol{\psi}_i^a$ that, similarly to $\boldsymbol{\omega_i}$, only affects the color branch of the model. We define the concatenation of shape and appearance latent codes as the object instance code $\boldsymbol{\psi}_i = (\boldsymbol{\psi}_i^s, \boldsymbol{\psi}_i^a)$ defined in the previous section.

Our resulting foreground object model is thus a conditional NeRF that takes the following form:
\begin{equation}
F^f \left(\bb{x}, \bb{d}, \left(\boldsymbol{\psi}_i^s,\boldsymbol{\psi}_i^a\right) \right) = G\left(D\left(\bb{x}, \boldsymbol{\psi}_i^s\right), \bb{d}, \boldsymbol{\psi}_i^a\right). \\
\end{equation}
We visualize our complete network architecture in Fig.~\ref{fig:model_architecture}.

\subsection{Loss Functions}

\paragraph{Photometric Loss} We apply the standard photometric L2 loss
\begin{equation}
    \ell_{\text{fg}} = \sum_{\bb{r} \in D_{\text{fg}}} \left\|\bb{C}(\bb{r}) - \bb{\bar{C}}(\bb{r}) \right\|_2^2\,,
\end{equation}
where $D_\text{fg} = \cup_{i=1}^N D_i$ and $\bb{\bar{C}}(\bb{r})$ is the ground truth RGB value for ray $\bb{r}$. Additionally, to ensure our background model learns the appropriate background geometry, we apply the same loss to background:
\begin{equation}
    \ell_{\text{bg}} = \sum_{\bb{r} \in D_{\text{bg}}} \left\|\bb{C}(\bb{r}) - \bb{\bar{C}}(\bb{r}) \right\|_2^2\,.
\end{equation}

\paragraph{Separation Regularization} The structure of our model (Figure \ref{fig:model_architecture}) allows us to disentagle foreground and background when the dataset satisfies the assumption of identical background geometry across different scenes. However, this separation doesn't naturally appear during the optimization, thus we apply a regularization loss to encourage it.
We consider the accumulated foreground density along a ray:
\begin{equation}
    A_f(\bb{r}) = \int_{t_n}^{t_f} T_f(t) \sigma_f(t) dt \,,
\end{equation}
where $T_f(t) = \exp\left( - \int_{t_n}^t \sigma_f(s) ds \right)$. We impose an L1 sparsity penalty $\ell_{\text{sparse}} = \| A_f \|_1$. This encourages $F^b$ to represent as much of the scene as possible. However, since the background model does not vary its density, $F^f$ is forced to represent any varying geometry of the scenes, which includes the objects.
Thus, the separation is not supervised, and occurs due to this sparsity prior and model structure.

While $\ell_{\textrm{sparse}}$ helps to separate foreground and background, it tends to encourage $F^f$ to pick up artifacts such as shadows induced by the object and lighting changes. These artificats typically manifest as haze-like artifacts, thus we suppress them with a beta prior loss function inspired by \cite{lombardi2019neural}:
\begin{equation}
    \ell_{\textrm{beta}} = (\alpha - 1) \log(A_f) + (\beta-1) \log(1 - A_f)
\end{equation}
In our experiments, we set $\alpha=3, \beta=2$, which biases $A_f$ towards 0, which is in line with our sparsity loss.

\paragraph{Deformation Regularization} Finally, we apply a simple L2 loss on the residual translation to penalize the deformation from being arbitrarily large:
\begin{equation}
    \ell_\text{warp} = \expec\left[ \| D(\bb{x}, \boldsymbol{\psi}_i^s) \|_2^2 \right]
\end{equation}
where the expectation is approximated with Monte Carlo sampling at the sampled points in the volume rendering procedure~\cite{mildenhall2020nerf}.

Our full loss function is described by: $\ell_\text{fg} + \lambda_\text{bg} \ell_\text{bg} + \lambda_\text{sparse} \ell_\text{sparse} + \lambda_\text{beta} \ell_\text{beta} + \lambda_\text{warp} \ell_\text{warp}$.

\section{Experiments}
\label{sec:experiments}

\subsection{Implementation Details}

Our architecture deviates slightly from the original NeRF paper. We add a density branch, and set the backbones of $F^f, F^b$ to 2 hidden layers with 256 units. The color and density branches have 8 layers of 128 hidden units each, with a skip connection at layer 5. Our deformation field MLP has 6 hidden layers with 128 units each and a skip connection at layer 4.
Following~\cite{park2020nerfies}, we use coarse and fine background and template models, but only one deformation field.
We jointly train all models together for 500k iterations using the same schedule as the original NeRF~\cite{mildenhall2020nerf} while using a batch size of 4096 rays on 4 V100 GPUs, which takes approximately 2 days.
We set $\lambda_\text{bg} = 1, \lambda_\text{sparse} = 1e{-3}, \lambda_\text{beta}=1e{-4}, \lambda_\text{warp}=1e{-5}$ except for ablations. We apply a top-$k$ schedule to $\ell_\text{beta}$ so that it focuses more on hard negatives after every 50k iterations.
We additionally apply the coarse-to-fine regularization scheme of~\cite{park2020nerfies} to all models for 50k iterations to prevent overfitting to high frequencies.
The latent code dimension is set to 64 for all experiments.
Finally, we use the same number of coarse and fine samples as NeRF, except for the \textsc{\small Glasses} dataset which uses 96 coarse samples to better capture thin structures.
More details can be found in the appendix.

\postsubsec
\subsection{Datasets}
\label{sec:datasets}
We use 3 datasets: \textsc{\small Cars} (synthetic), \textsc{\small Glasses} (real, controlled lab capture), \textsc{\small Cups} (real, hand-held capture) as described below:

\paragraph{\textsc{\small Cars}:} We render 100 cars from ShapeNet~\cite{shapenet2015} using Blender~\cite{blender} at a 128x128 resolution. We randomly sample an elevation, then sample viewpoints by linearly spaced azimuth points. The cars are rendered against a gray canvas; the background can be assumed to be an infinitely long gray plane without any appearance variation.

\paragraph{\textsc{\small Glasses}:} We use 60 glasses from the dataset of~\cite{martinbrualla2020gelato}, which consists of a lab-capture setup where eyeglasses are placed on a mannequin.
The images exhibit real-world phenomena such as lighting variations and shadows, making it more challenging to learn compared to synthetic data. Additionally, glasses are challenging to model due to thin geometric structures that they possess. 
We use images cropped to 512x512. 
The mannequin setup has a backlight which was used to capture additional images in order to extract foreground alpha mattes~\cite{martinbrualla2020gelato}; we arbitrarily threshold these at .5 to obtain segmentation masks. However, we show that the backlit-captured segmentation masks can be quite noisy, and our learned segmentations are much cleaner and more accurate without the need for the backlight nor additional images. Note that we use these masks purely for evaluation, not for training. We do not use the backlit images either.

\begin{table}
\centering
\resizebox{\linewidth}{!}{\begin{tabular}{c||c:cccc}
\toprule
  & \multicolumn{5}{c}{\textsc{\small Glasses}} \\
\cline{2-6}  $\ell_\text{sparse}$ & \imagefidelitymetric{FID} $\downarrow$ & \heldoutmetric{PSNR}$\uparrow$ & \heldoutmetric{SSIM}$\uparrow$ & \heldoutmetric{LPIPS}$\downarrow$ & \heldoutmetric{IoU}$\uparrow$ \\ \hline
 \xmark  & 40.00 & 36.98 & .9420 & .0970 & .0410 \\
 \cmark  & 39.56 & 36.26 & .9402 & .0968 & .5796 \\
\midrule
  & \multicolumn{5}{c}{\textsc{\small Cups}} \\
\cline{2-6} \xmark & 102.8 & 25.23 &.9441  & .0634 & .1403 \\
 \cmark & 101.8 & 25.20 & .9440 & .0630 & .8560 \\
\bottomrule
\end{tabular}
}
\caption{Loss function ablation with our full model. For qualitative results of the effect of $\ell_\text{beta}$, see Figure~\ref{fig:loss_function_ablation}.}
\label{table:loss_function_ablation}
\end{table}

\removed{

\begin{table}
\centering
\resizebox{\linewidth}{!}{\begin{tabular}{cc||c:cccc}
\toprule
  &  & \multicolumn{5}{c}{\texttt{Glasses}} \\
\cline{3-7}  $\ell_\text{sparse}$ & $\ell_\text{beta}$ & \imagefidelitymetric{FID} $\downarrow$ & \heldoutmetric{PSNR}$\uparrow$ & \heldoutmetric{SSIM}$\uparrow$ & \heldoutmetric{LPIPS}$\downarrow$ & \heldoutmetric{IoU}$\uparrow$ \\ \hline
 \xmark & \xmark & 40.00 & 36.98 & .9420 & .0970 & .0410 \\
 \xmark & \cmark & 40.07 & 37.03 & .9428 & .0961 & .0315 \\
 \cmark & \xmark & 41.01 & 37.03 & .9426 & .0967 & .6124 \\
 \cmark & \cmark & 39.56 & 36.26 & .9402 & .0968 & .5796 \\
\midrule
  &  & \multicolumn{5}{c}{\texttt{Cups}} \\
\cline{3-7} \xmark & \xmark & 102.8 & 25.23 &.9441  & .0634 & .1403 \\
 \xmark & \cmark & 104.7 & 25.12 & .9436 & .0644 & .1292 \\
 \cmark & \xmark & 98.89 & 25.23 & .9442 & .0629 & .8578 \\
 \cmark & \cmark & 101.8 & 25.20 & .9440 & .0630 & .8560 \\
\bottomrule
\end{tabular}
}
\caption{Loss function ablation with our full model. \chrisnote{ metrics aren't really affected by the losses, but $\ell_\text{sparse}$ is clearly needed to separate. Should we somehow condense this table?}}
\label{table:loss_function_ablation}
\end{table}

}
\begin{figure}[t]
\centering
\includegraphics[width=\linewidth]{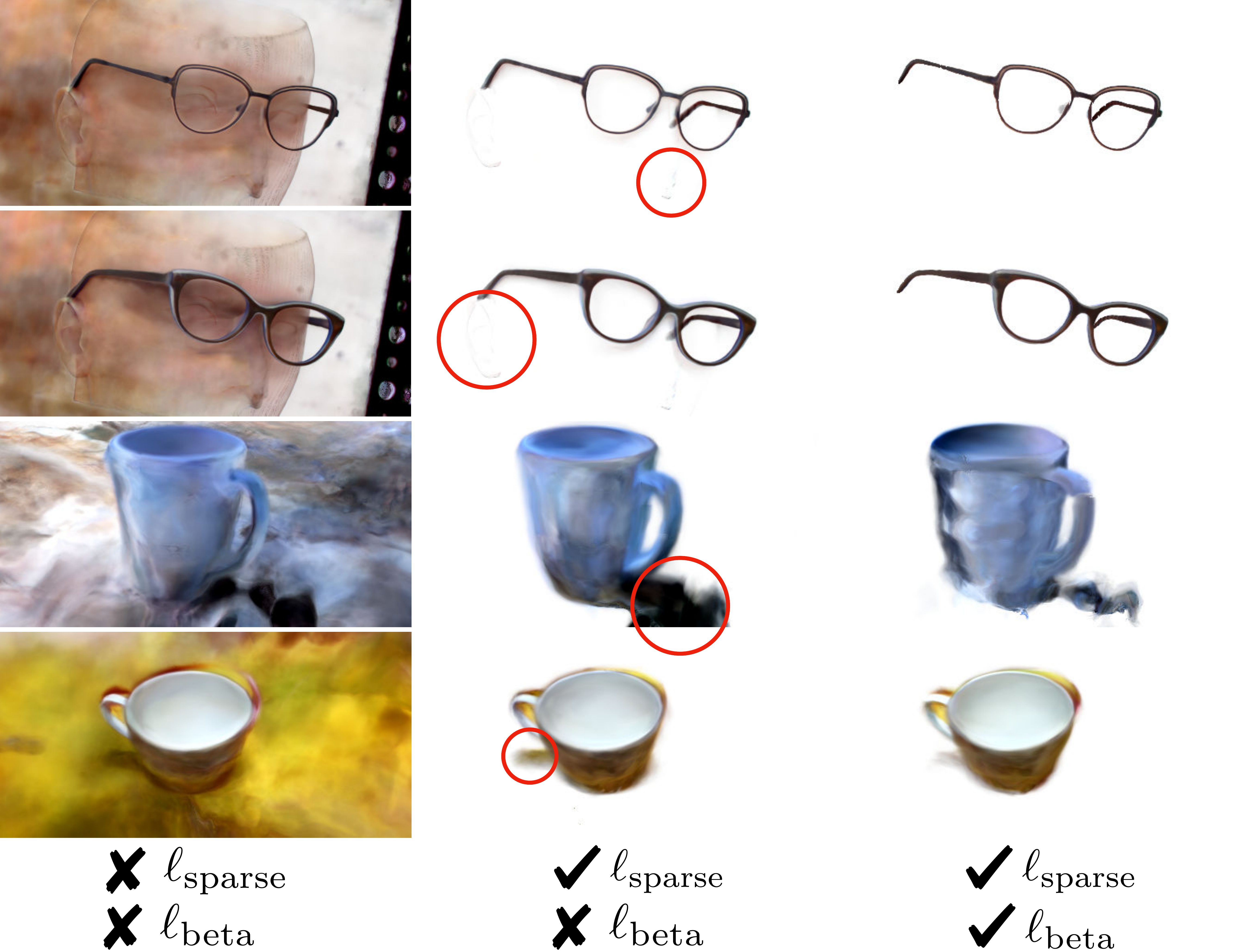}
\caption{Foreground renderings of ablated models. 
Without $\ell_\text{sparse}$ (left), the model fails to separate the background from the foreground.
Lingering artifacts, such as faint halos around mannequin silhouette (middle, circled in red), are further suppressed with $\ell_\text{beta}$ (right). Best viewed zoomed in.}
\label{fig:loss_function_ablation}
\postfigure
\end{figure}

\paragraph{\textsc{\small Cups}:} We build this dataset from the cups class from Objectron~\cite{objectron2020}. Objectron consists of casually-captured video clips in-the-wild which are accompanied by camera poses and manually annotated bounding boxes. In each video, the camera moves around the object, capturing it from different angles. 
While the cups are filmed while resting on planar surfaces, they typically have varying background geometries in the scene, which violates our assumption. Thus, we manually selected 100 videos in which the 2D projection of the bounding box annotation covers a planar region. We only use pixels within this region for both training and evaluation. We crop and resize the frames to 512x512, and use the coordinate frames centered at the object, giving us alignment between the instances. To obtain segmentation masks for evaluation, we use Mask R-CNN~\cite{he2017mask} with a SpineNet96 backbone~\cite{du2020spinenet}. In order to ground the background to a planar geometry, we manually capture a hardwood floor with textured papers for $D_\text{bg}$ (see Figure~\ref{fig:dataset_setup}).

All datasets use 50 images per object instance (including background) for training, and 20 images per instance for heldout validation. The near and far bounds $t_n, t_f$ are selected for each dataset manually, while for \textsc{\small Cups} we use per-instance values since the camera can be at a different distance from the object in each video.

\begin{figure}[t]
\centering
\includegraphics[width=\linewidth]{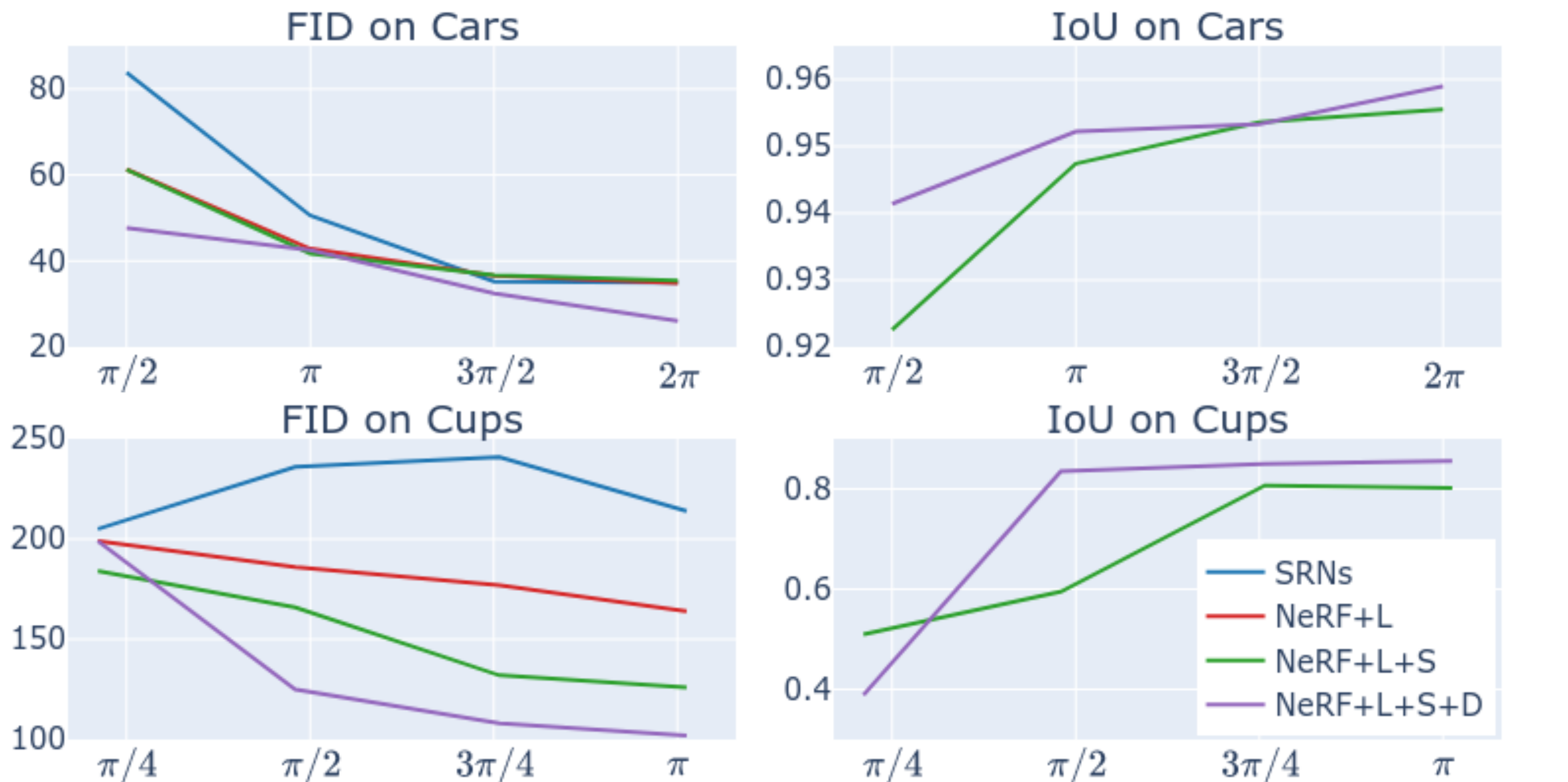}
\caption{Comparison of methods under limited viewpoint ranges. The x-axis is the limit of the azimuth range per-instance during training. See text for how this is set.
}
\label{fig:limited_viewpoints}
\postfigure
\end{figure}

\postsubsec
\subsection{Baselines}

We establish a baseline that simply adds a latent code to NeRF (\nerfl{}). We implement this by essentially using only $F^f$ for the composite rendering and conditioning it with $\boldsymbol{\psi}_i^s$ on its density branch (see Figure~\ref{fig:model_architecture}). 
Additionally, we train Scene Representation Networks (SRNs)~\cite{sitzmann2019srns}, a recent state-of-the-art method for object category modelling, as a baseline. We trained SRNs for 300k iterations with a batch size of 8, which took approximately 3 days. The latent embedding is set to 64.

Since SRNs and \nerfl{} cannot separate foreground and background, we also compare against a third baseline that uses our architecture without the deformation field. For this model (\nerfls{}, S for separation), we take \nerfl{} and add the background model. Comparisons with this baseline allows us to evaluate the efficacy of the deformation field in modelling object categories.
Note that \nerflsd{} (D for deformation) = FiG-NeRF.

\postsubsec
\subsection{Metrics}

To evaluate how well a model learns an object category, we test its ability to synthesize novel viewpoints and its ability to interpolate instances. 
We report PSNR, SSIM, and LPIPS~\cite{zhang2018unreasonable} for view synthesis and Frechet Inception Distance (FID)~\cite{heusel2017gans} for instance interpolation. 
Note that we sample random instance interpolation points in the latent space and fix these for all models for a more fair comparison. Lastly, we show Intersection over Union (IoU) with respect to the obtained segmentation masks on heldout viewpoints to evaluate the foreground/background separation. Section~\ref{sec:datasets} describes the segmentation reference methods.

\postsubsec
\subsection{Results}

\begin{figure}[t]
    \centering
    \includegraphics[width=\linewidth]{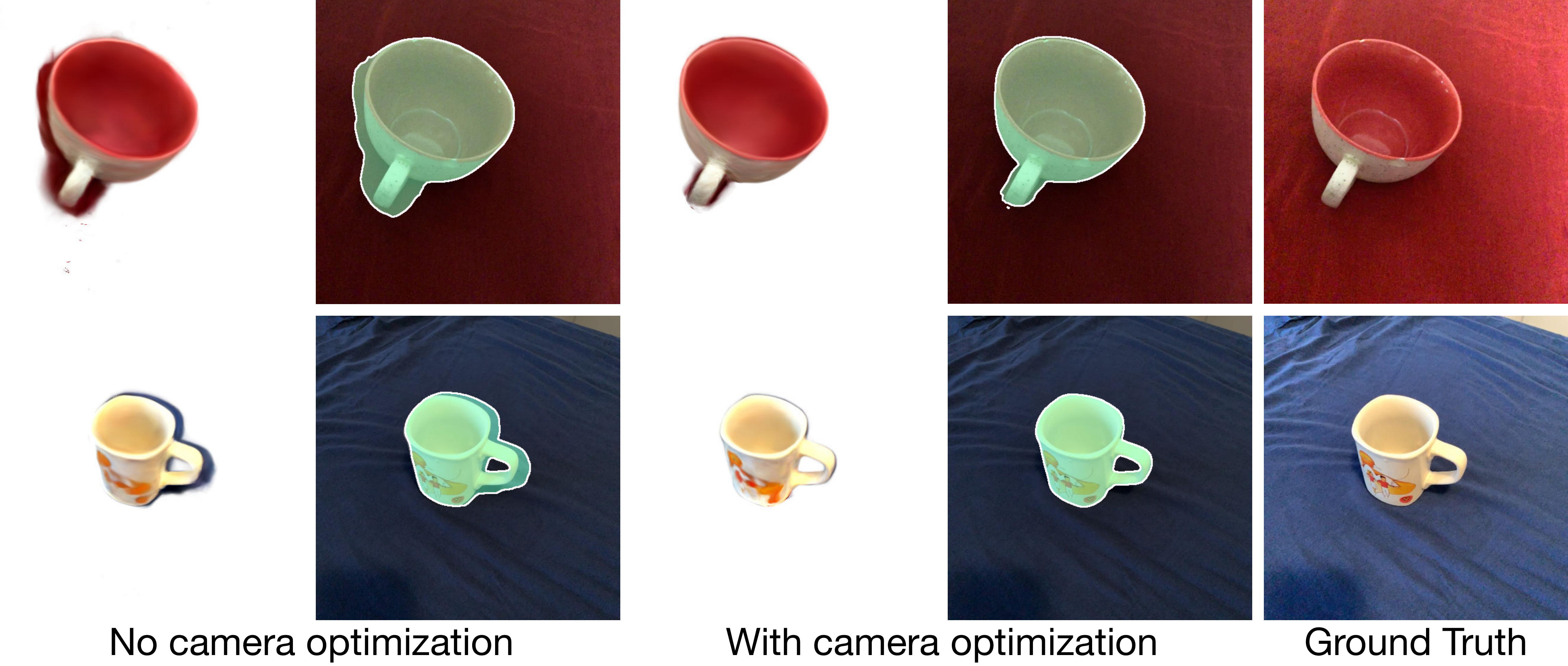}
    \caption{Optimizing camera positions jointly with FiG-NeRF leads to better geometry on \textsc{\small cups}. Columns 1 and 2 show the foreground rendering and segmentation mask from FiG-NeRF without camera optimization, and columns 3 and 4 show FiG-NeRF with camera optimization.}
    \label{fig:camera_optimization}
    \postfigure
\end{figure}

\subsubsection{Model Comparison}

We compare our proposed method with the baselines on all datasets in Table~\ref{table:model_comparison}. On the clean synthetic \textsc{\small Cars} dataset, all models and baselines perform well. The NeRF-based models (NeRF+*) outperform SRNs on the heldout metrics including PSNR/SSIM/LPIPS. While the NeRF-based models show similar performance on the heldout metrics, \nerflsd{} (FiG-NeRF) significantly outperforms all other baselines on FID, demonstrating its ability to interpolate between instances and model the object category.

On the \textsc{\small Glasses} dataset, the region of the image occupied by the glasses themselves is fairly small. We believe this to be the reason that the results are very similar for the NeRF-based models. However, the models with separation clearly outperform \nerfl{} on FID. While this baseline only has half of the parameters of our model, it cannot decompose the scenes into foreground and background, and these results suggest that having a dedicated model to foreground aids performance when interpolating instances. 
Additionally, \nerflsd{} (FiG-NeRF) shows a slight performance increase compared to the non-deformable baseline on the heldout metrics. 
Please see the project website for qualitative differences between \nerfls{} and \nerflsd{} (FiG-NeRF).
Lastly, note that SRNs failed on this dataset since it cannot handle varying camera intrinsics.

For \textsc{\small Cups}, SRNs perform poorly for this complex real-world dataset, indicated by a 30\% decrease in PSNR from the NeRF-based models. \nerfls{} performs very similarly to \nerflsd{} (FiG-NeRF) on PSNR/SSIM/LPIPS, but is significantly outperformed on FID and IoU. As this dataset is the most challenging, this suggests that the deformation field notably aids in learning how to interpolate instances, and also the geometry. \nerfl{} has much more trouble interpolating instances as shown in FID. These results suggest that our proposed model is adept not just in synthesizing new views, but also at interpolating between instances. 

\begin{figure}[t]
\centering
\includegraphics[width=\linewidth]{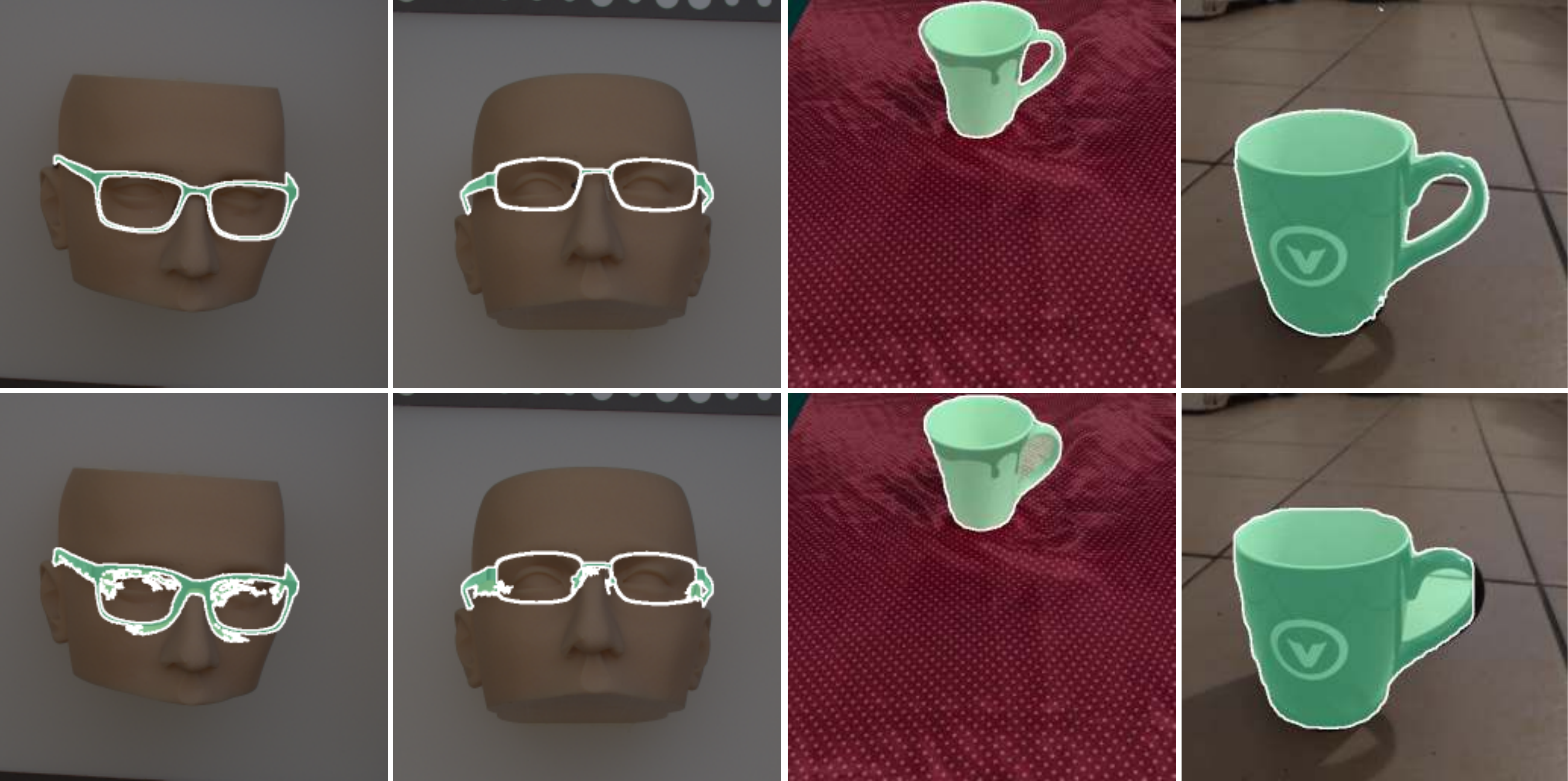}
\caption{Segmentation Comparison. The segmentations in the $1^\text{st}$ row are produced by our model, and the $2^\text{nd}$ row is produced by the backlit matte extraction~\cite{martinbrualla2020gelato} and Mask R-CNN~\cite{he2017mask}. Our method learns more accurate and crisp segmentation masks without supervision.
}
\label{fig:segmentation_comparison}
\postfigure
\end{figure}

\begin{figure*}[t]
    \centering
    \begin{subfigure}[b]{0.49\textwidth}
        \centering
        \includegraphics[width=\textwidth]{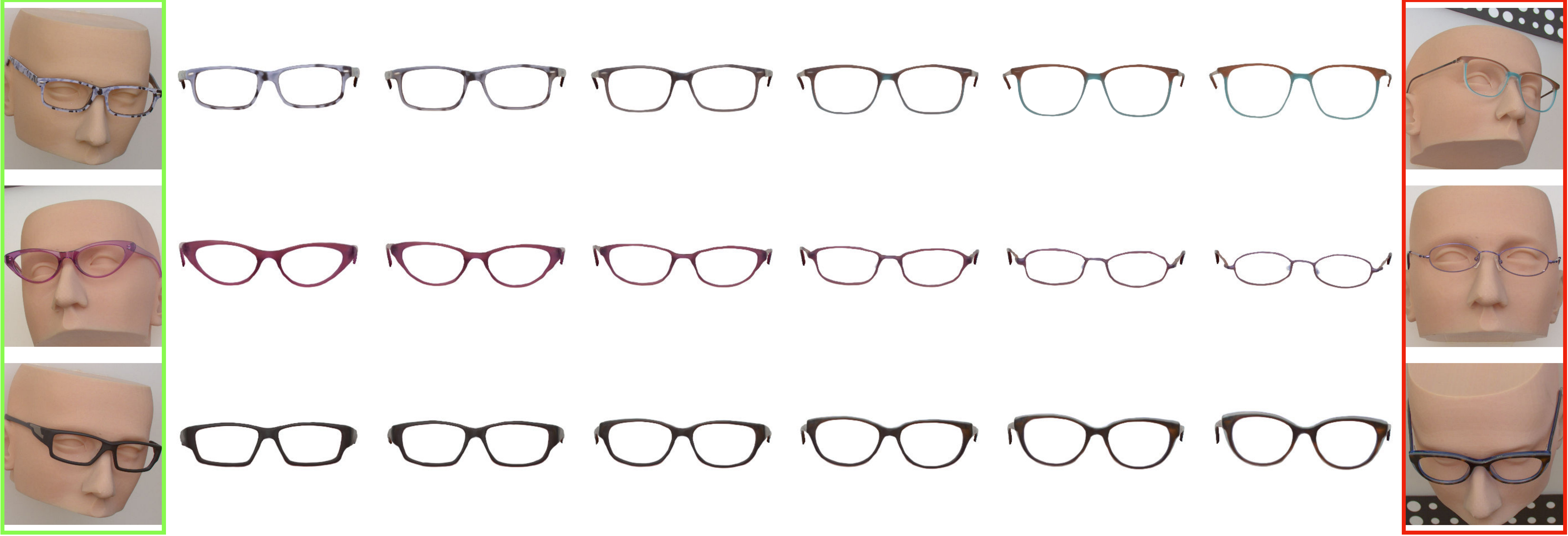}
        \label{subfig:glasses_interpolation}
    \end{subfigure}
    \hfill
    \begin{subfigure}[b]{0.49\textwidth}
        \centering
        \includegraphics[width=\linewidth]{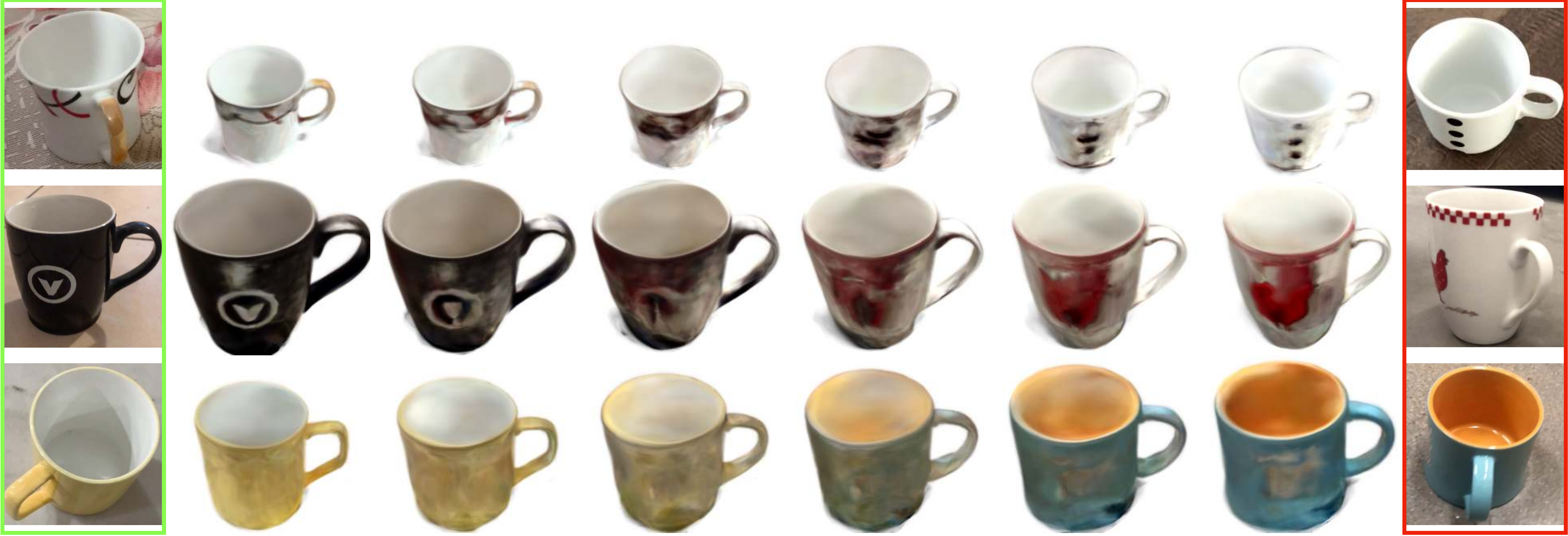}
        \label{subfig:cups_interpolation}
    \end{subfigure}

    \caption{Shape and color interpolations on \textsc{\small Glasses} and \textsc{\small Cups}. We show cropped training images on the left ({green} box) and right (\textcolor{red}{red} box) columns. In the middle, we show interpolations rendered from the foreground model.
    Best viewed zoomed in.}
    \label{fig:color_and_shape_interpolation}
    \postfigure
\end{figure*}

\postsubsubsec
\subsubsection{Loss Function Ablation}

In this section, we demonstrate that the network structure alone is not enough for decomposition, and that $\ell_\text{sparse}$ is required to induce the separation.
We perform these experiments with our full model on the \textsc{\small Glasses} and \textsc{\small Cups} and show results in Table~\ref{table:loss_function_ablation}. On both datasets, the IoU shows that it is clear that $\ell_\text{sparse}$ is needed to learn the separation, and this leaves the composite rendering relatively unaffected as evidenced by FID, PSNR, SSIM, and LPIPS.
Additionally, $\ell_\text{beta}$ is also useful in learning in the separation. It aids in suppressing artifacts such as shadows and lighting variation from appearing in the foreground model, which results in a clear delineation between foreground and background. These effects are too subtle to influence the quantitative metrics, thus we omit these numbers from Table~\ref{table:loss_function_ablation}. However, we demonstrate these effects qualitatively in Figure~\ref{fig:loss_function_ablation}.

\postsubsubsec
\subsubsection{Robustness to Limited Viewpoints}

Having a large range of viewpoints for each object helps in learning object geometries. However, in many real-world settings, this may not hold. For example, most of the videos in \textsc{\small Cups} only show half of the cup. We show that our deformable formulation fares better in low viewpoint range regimes. We test this with \textsc{\small Cars} and \textsc{\small Cups} by creating datasets from them that have 4 levels of varying azimuth ranges, denoted by $\Delta \theta$. For each instance in \textsc{\small Cars}, we render from viewpoints that are constrained with $\Delta \theta \in \{.5\pi, \pi, 1.5\pi, 2\pi\}$. For \textsc{\small Cups}, we do not have such control over the data. Instead, for each video capture we sort all viewpoints by their azimuth angle and keep only the first 25\%, 50\%, 75\% and 100\% of the viewpoints. This results in mean and standard deviations of $\Delta \theta \sim \{ (0.6, 0.5), (1.5, 0.6), (2.4, 0.8), (3.2, 1.2) \}$ for each dataset, where $X \sim (\mu, \sigma)$ means the random variable $X$ is distributed with mean $\mu$ and standard deviation $\sigma$. 

We evaluate instance interpolation and geometry reasoning in Figure~\ref{fig:limited_viewpoints} with FID and IoU (as a proxy to geometry).
Obviously, as the azimuth range decreases, so does the performance of each model. However, on the synthetic \textsc{\small Cars} dataset, our proposed model fares better on both metrics in these low viewpoint range settings. Additionally, on \textsc{\small Cups}, our model still performs well as $\Delta\theta$ reaches 1.5 on average, and still retains an IoU of 0.835 while \nerfls{} drops to 0.595. These results show that our formulation better captures the nature of object categories, and is desirable to deploy in these viewpoint constrained settings.

\subsubsection{Optimizing Camera Poses}

\begin{table}[h]
\centering
\resizebox{\linewidth}{!}{\begin{tabular}{c|cc|c|cc}
\toprule
offsets & \imagefidelitymetric{FID} $\downarrow$ & IoU$\uparrow$ & offsets & \imagefidelitymetric{FID} $\downarrow$ & IoU$\uparrow$ \\ \hline
\xmark & \tablesecond 106.4 & \tablesecond 0.8543 & \cmark & \tablefirst 85.67 & \tablefirst 0.8967 \\ 
\bottomrule
\end{tabular}
}
\caption{Due to inaccurate camera poses in Objectron~\cite{objectron2020}, jointly optimizing FiG-NeRF and camera extrinsics leads to better object modelling on \textsc{\small cups}. Note that IoU is evaluated on training set.}
\label{table:camera_optimization}
\end{table}

Camera poses from Objectron~\cite{objectron2020} are obtained via visual-inertial odometry, which are prone to drift and can result in inaccurate camera estimates. For the instances in \textsc{\small cups}, this causes significant jitter in the camera poses between consecutive frames of the same video, thus the instance appears to move slightly within the 3D volume. This results in the model learning to put mass in an envelope that contains all of these offsets, as seen in Figure~\ref{fig:camera_optimization}. We address this issue by optimizing the camera extrinsics during training. In particular, we leave the camera parameters fixed for the first 50k iterations of training to let the FiG-NeRF separate foreground from background, and then learn offsets to each image's camera extrinsics for the rest of the training procedure. Additionally, we set $\lambda_\text{warp}=0$ since object scale is ambiguous as we optimize cameras. In Table~\ref{table:camera_optimization}, we see that training the cups model with the camera optimization results in much better FID and IoU, indicating better interpolations and geometry. Note that because we optimized camera poses, IoU is computed on the training set since we no longer have ground truth cameras for the heldout views.

\postsubsubsec
\subsubsection{Segmentation Results}
\label{subsubsec:segmentation_results}

Since our reference segmentations are noisy, we qualitatively compare them with our learned segmentations in Figure~\ref{fig:segmentation_comparison}. 
We show that our model can effectively capture thin structures in the foreground component on \textsc{\small Glasses}. Additionally, on \textsc{\small Cups} our model frequently learns to correctly label the free space in the handle of the cup as background.
Unfortunately, we do not have access to segmentations that are accurate enough to reflect this quantitatively. 
These results are learned by leveraging the model structure and sparsity priors without any supervision for the separation.

Because our foreground component learns a representation of the object from multiple views, in settings where the background occludes the foreground (e.g. the mannequin head in \textsc{\small glasses} can occlude the glasses temples from certain viewpoints), we can easily compute an amodal segmentation by thresholding the accumulated foreground density $A_f$. 
We show examples of this on the project website.

\subsubsection{Interpolations}

\begin{figure*}[t]
    \centering
    \begin{subfigure}[b]{0.49\textwidth}
        \centering
        \includegraphics[width=\textwidth]{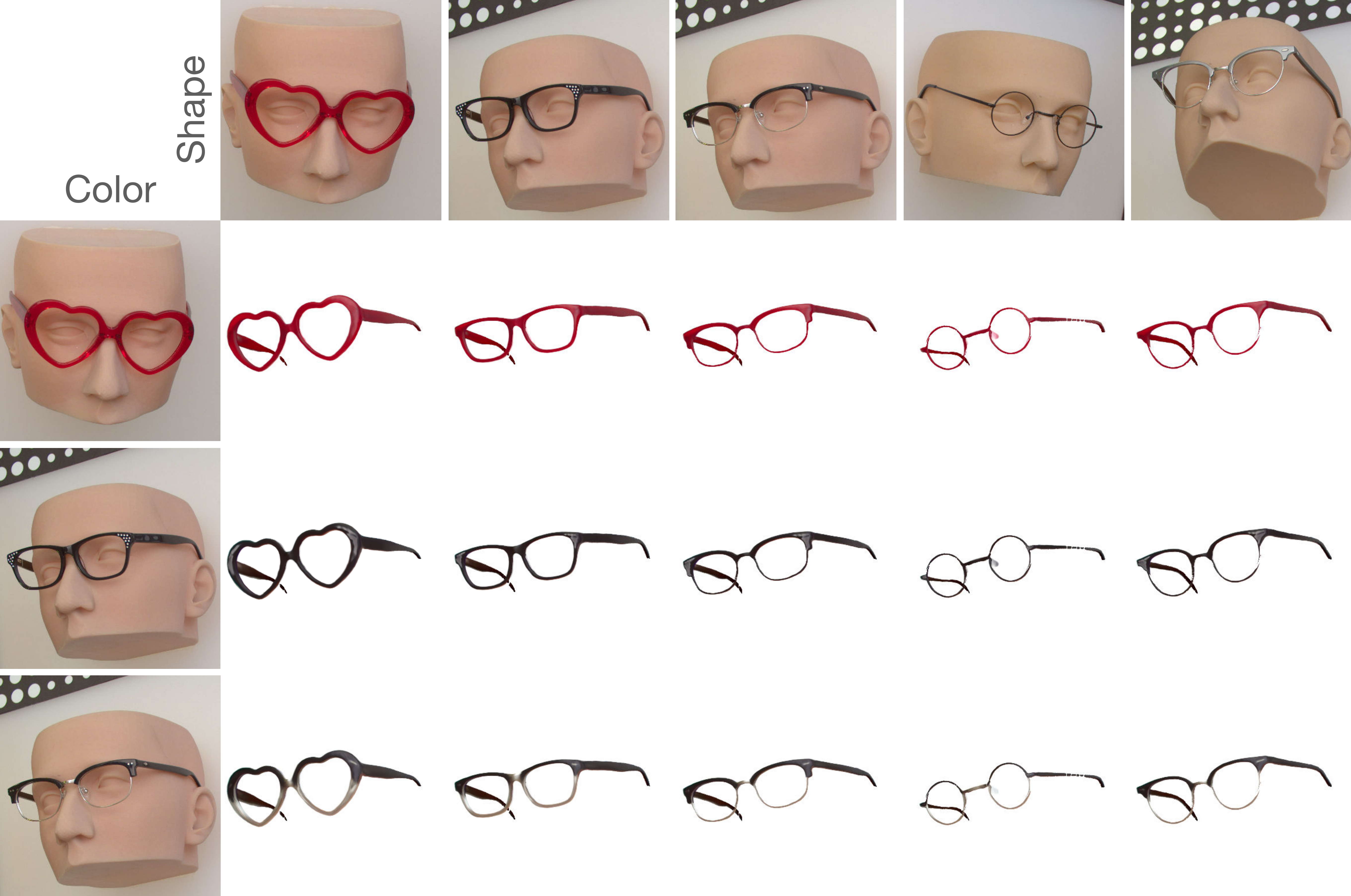}
        \label{subfig:glasses_fixed_color_or_density}
    \end{subfigure}
    \hfill
    \begin{subfigure}[b]{0.49\textwidth}
        \centering
        \includegraphics[width=\linewidth]{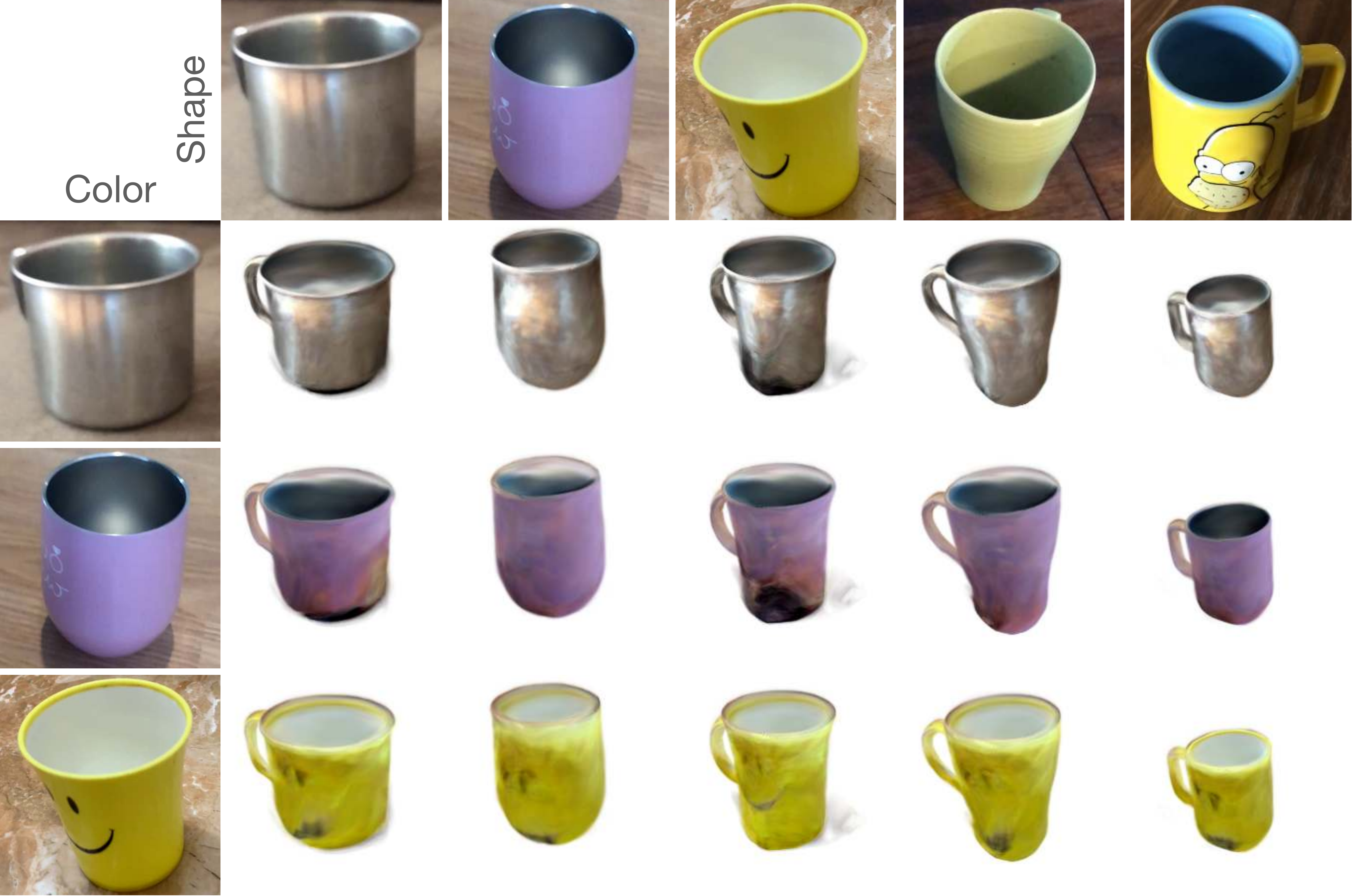}
        \label{subfig:cups_fixed_color_or_density}
    \end{subfigure}
        
    \caption{We show examples of rendering \textsc{\small Glasses} and \textsc{\small Cups} while keeping either the shape or color fixed. The left column and top row shows training images, while the middle shows foreground renderings.
    Best viewed in color and zoomed in.}
\label{fig:fixed_color_or_density}
\postfigure
\end{figure*}

We visualize the ability of our model to interpolate between instances in Figure~\ref{fig:color_and_shape_interpolation} for \textsc{\small Glasses} and \textsc{\small Cups}. In the rendered foreground, we demonstrate smooth interpolations between the instances. The midpoints of the interpolations give plausible objects. For example, in the $2^\text{nd}$ glasses interpolation, the midpoint shows a generated pair of glasses that exhibits the shape of the left pair (in the green box) while the frame has the thinness of the right pair (in the red box). 
In the $3^\text{rd}$ cup interpolation, we see the cup size smoothly increasing from left to right. Additionally, the midpoint shows a curved handle which is a feature of the cup on the right, but the handle size reflects the cup on the left.

Since color and shape are disentangled by our network structure, we can fix one while interpolating the other. In Figure~\ref{fig:fixed_color_or_density}, we demonstrate this on \textsc{\small Glasses} and \textsc{\small Cups}. 
Our model is able to adapt the texture to the other shapes. In particular, row 3 of \textsc{\small Glasses} shows multi-colored glasses, where the color boundaries can be evidently seen in the thicker glasses.
Additionally, the texture of the black glasses with the beads in the corner ($2^\text{nd}$ row) 
is successfully transferred to all of the other glasses shapes.
On \textsc{\small Cups}, we successfully transfer the metal and purple textures to the other cups. 
Note that the $2^\text{nd}$ cup geometry has no handle.
Our model struggles to transfer the texture of the yellow cup to the other geometries, however the outline of the smiley face is visible.

\postsubsubsec
\subsubsection{Failure Cases}

\removed{
\begin{figure}[t]
\centering
\includegraphics[width=\linewidth]{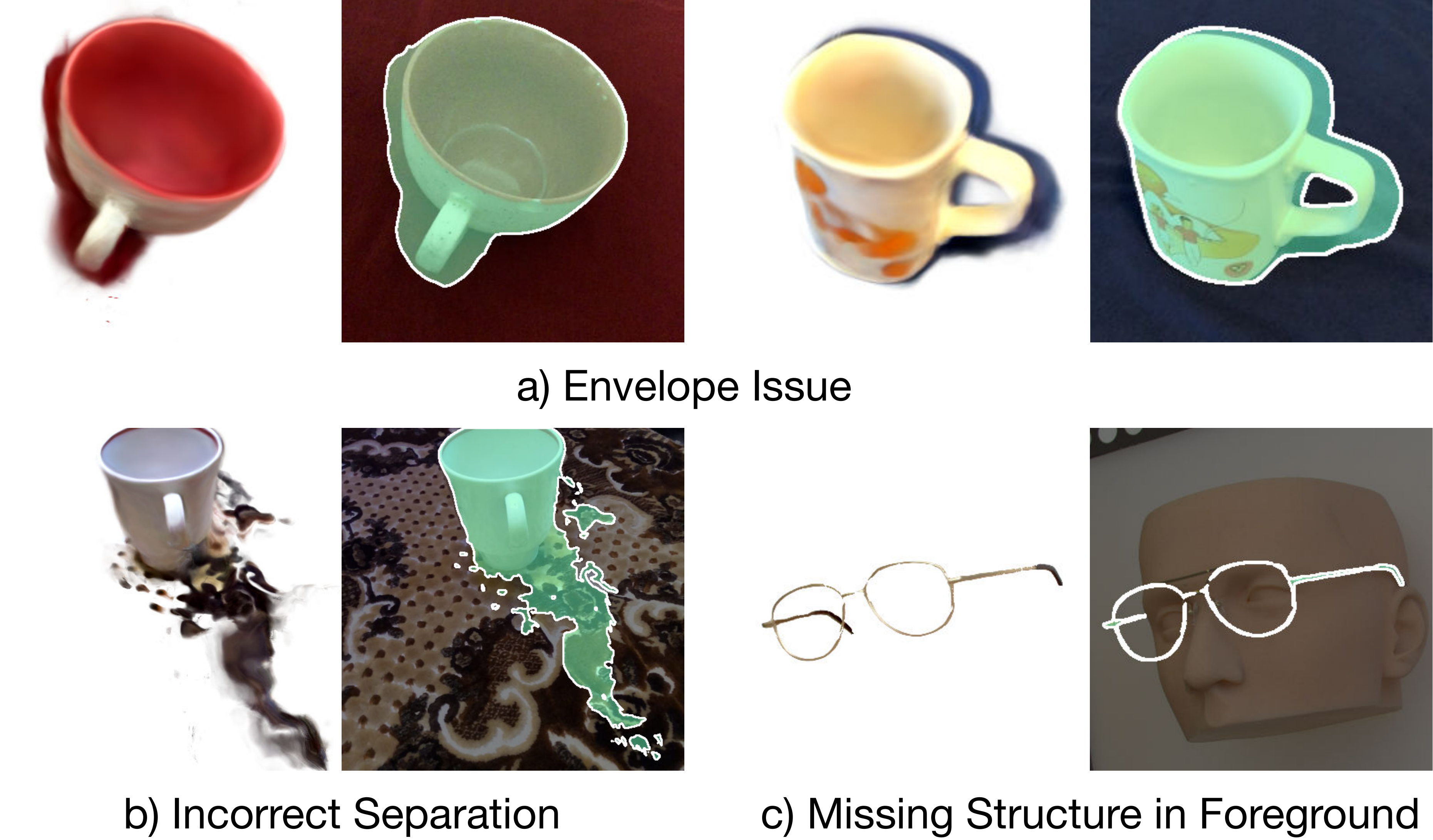}
\caption{We illustrate 3 failure cases. See text for discussion and analysis.}
\label{fig:failure_cases}
\end{figure}
}

\begin{figure}[t]
    
    \begin{subfigure}[b]{0.49\linewidth}
        \centering
        \includegraphics[width=\linewidth]{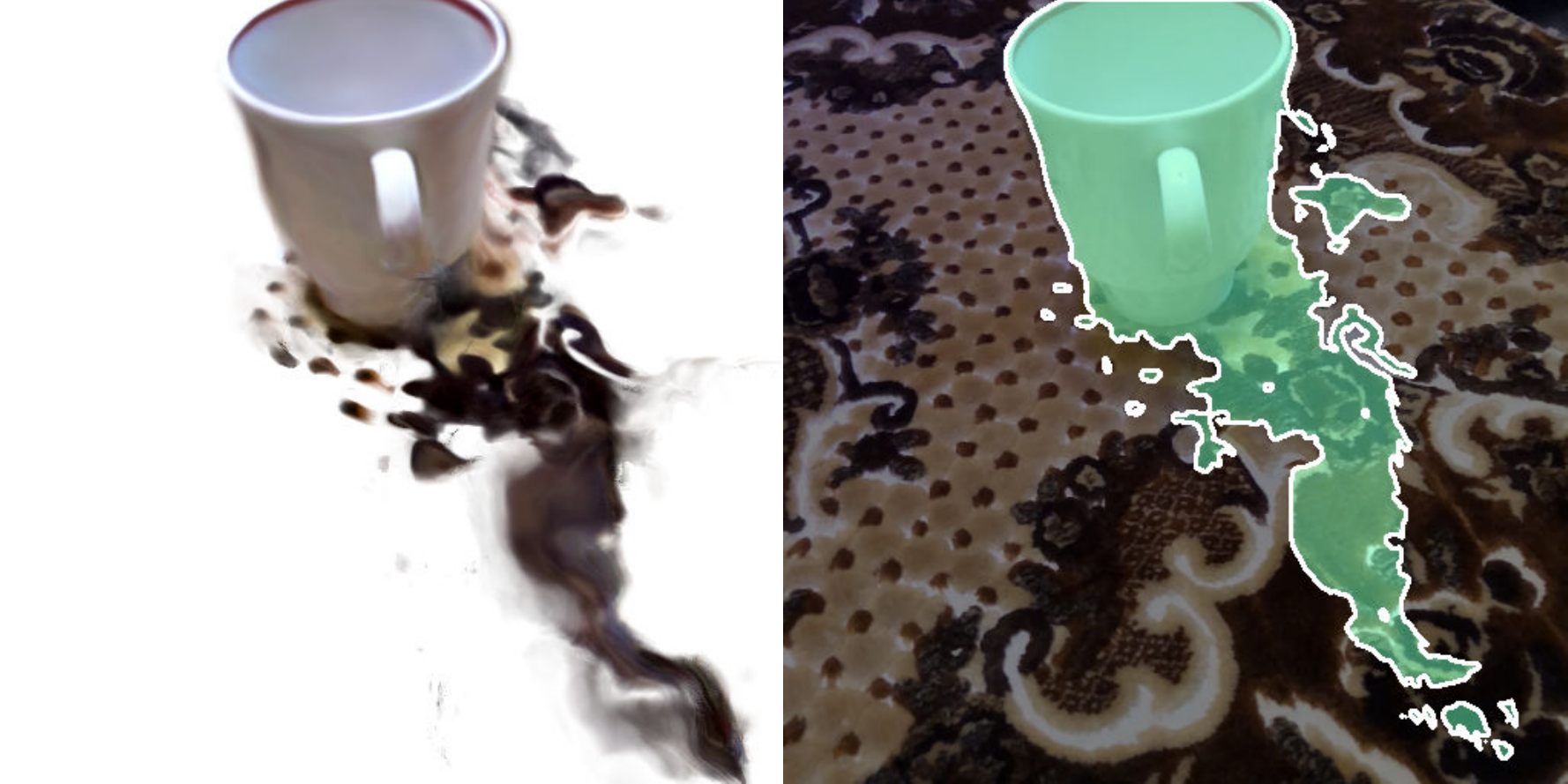}
    \end{subfigure}
    \hfill
    \begin{subfigure}[b]{0.49\linewidth}
        \centering
        \includegraphics[width=\linewidth]{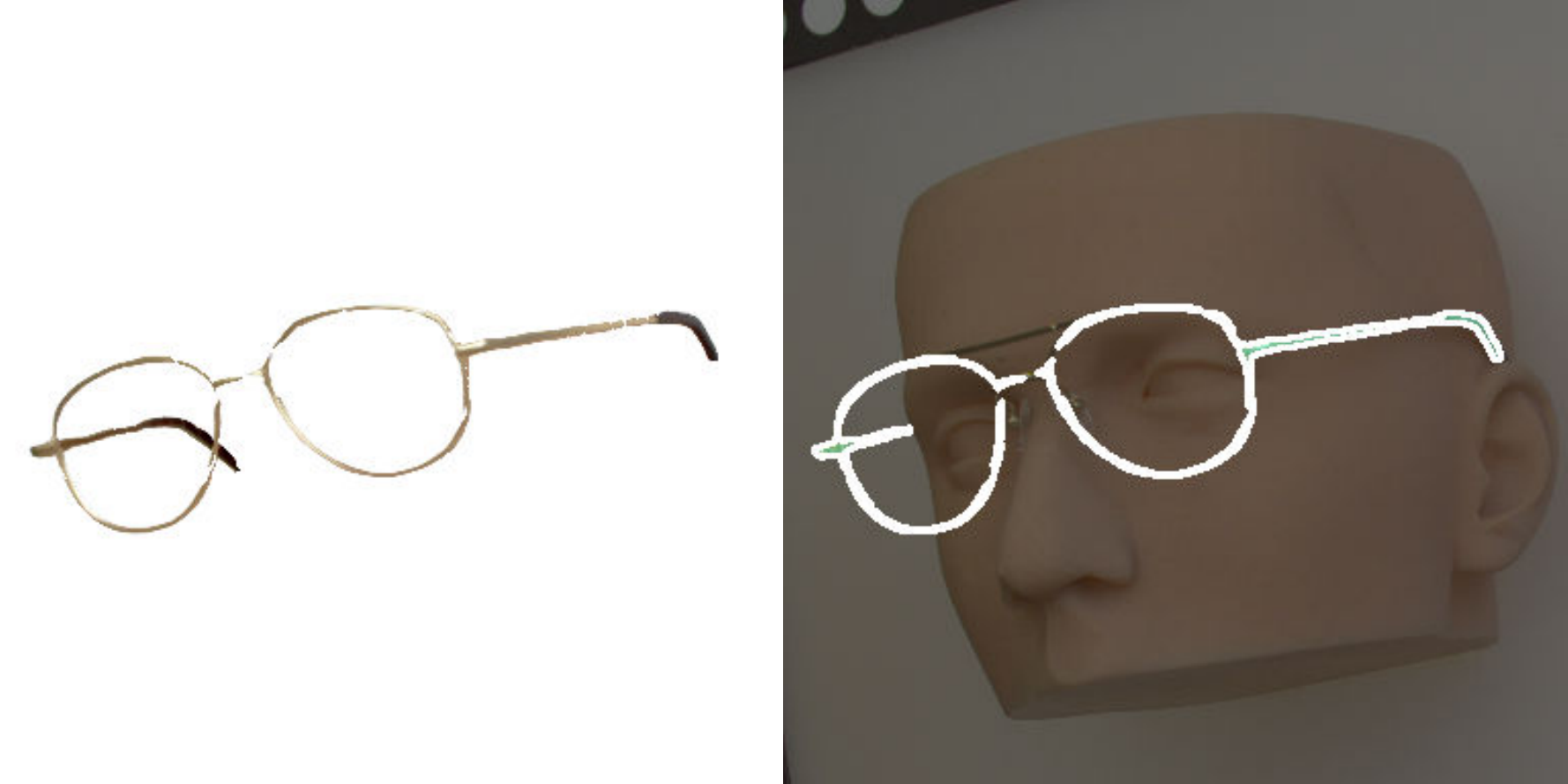}
    \end{subfigure}
    \caption{We show 2 failure modes. See text for discussion.}
    \label{fig:failure_cases}
    \postfigure
\end{figure}

We discuss two failure cases in Figure~\ref{fig:failure_cases}, showing foreground renderings and segmentation masks. 
First, Figure~\ref{fig:failure_cases} (left) shows that if the background has a complicated texture such as a quilt pattern, it can cause some of the background to leak into the foreground. Secondly, if the object has a similar color as background, $\ell_\text{sparse}$ can push this into the background model to achieve lower loss. In Figure~\ref{fig:failure_cases} (right), the foreground did not capture the top bridge of the glasses.

\postsubsec

\section{Conclusion and Future Work}

We have demonstrated a 2-component Deformable Neural Radiance Field model, FiG-NeRF, for jointly modeling object categories in 3D whilst separating them from backgrounds of variable appearance. We applied this to modeling object categories from handheld captures. Our results suggest that modeling 3D objects ``in-the-wild” is a promising direction for vision research. Future work might improve on our method by handling a larger diversity of backgrounds, and robustness to geometric errors in capture. 



\appendix

\section{Implementation Details}

\subsection{$\ell_\text{beta}$ Schedule}

We apply a top-$k$ schedule to $\ell_\text{beta}$ so that it focuses more on hard negatives. We apply the following schedule: [(50k, 0.0), (50k, 0.5), (50k, 0.25), (50k, 0.1), ($\infty$, 0.05)], where each item in the schedule $(T,k)$ means we apply $\ell_\text{beta}$ only to the top $k$ percentage of pixels with the highest loss for $T$ iterations. Additionally, since the Beta distribution density function has gradients approaching infinite magnitude around $\{0, 1\}$, we clip the input between $[1e{-4}, 1 - 1e{-4}]$ and contract (scale) the input to this loss to $[-0.4, 0.4]$ such that there are no gradients when the input is very close to $\{0,1\}$, and the maximum magnitude of the gradient isn't too large.

\subsection{Random Density Perturbation}

Additionally, we add some randomness to the initial portion of the training procedure in order to help encourage the separation. We perturb $\sigma_f$ and $\sigma_b$ (the foreground and background volume densities, respectively) with a $N(0,1)$ random variable for the first 50k iterations of training. This helps the training process avoid local minima where either the foreground or background learn the entire scene, and the other learns nothing.

\subsection{Integral Approximations}

We follow~\cite{mildenhall2020nerf, martinbrualla2020nerfw} and approximate the integrals with numerical quadrature. Let $\{t_k\}_{k=1}^K$ be samples along the ray $\bb{r}$ between the near/far bounds $t_n, t_f$ such that $t_1 < t_2 \ldots < t_K$. We define $\delta_k = t_{k+1} - t_k$. To approximate $\bb{C}(\bb{r})$ (Eq. (2) in main paper) with $\hat{\bb{C}}(\bb{r})$,
\begin{align}
\begin{split}
    \hat{\bb{C}}(\bb{r}) = \sum_{k=1}^K \hat{T}(k) \bigg( &\alpha(\sigma_f(t_k)\delta_k) \bb{c}_f(t_k) +\\ 
    &\alpha(\sigma_b(t_k)\delta_k) \bb{c}_b(t_k)\bigg)\,,
\end{split}
\end{align}
where $\hat{T}(k) = \exp\left( -\sum \limits_{j=1}^{k-1} \Big( \sigma_f(t_j) + \sigma_b(f_j) \Big) \delta_j \right)$ and $\alpha(x) = 1 - \exp(-x)$.

Additionally, to apply $\ell_\text{sparse}$, we approximate $A_f(\bb{r})$ (Eq. (6) in the main paper) with $\hat{A}_f(\bb{r})$ by computing
\begin{align}
\begin{split}
    \hat{A}(\bb{r}) = \sum_{k=1}^K \hat{T}_f(k) \alpha \big( \sigma_f(t_k)\delta_k \big)
\end{split}
\end{align}
where $\hat{T}_f(k) = \exp\left( -\sum \limits_{j=1}^{k-1}  \sigma_f(t_j) \delta_j \right)$.

\subsection{Positional Encodings}

For the background model and foreground template, we use a positional encoding with 10 frequencies, as in the original NeRF~\cite{mildenhall2020nerf}. For the deformation field, we use 10 frequencies for the spatial encoding on \textsc{\small cars} and \textsc{\small glasses} since they can exhibit high frequency geometry such as spoilers, side mirrors, and thin frames, while we use 4 frequencies for \textsc{\small cups} since they typically do not exhibit high frequency geometry. Our viewing directions use a positional encoding with 4 frequencies.

\section{Dataset Details}

\subsection{ShapeNet}

For this dataset, we removed $\omega_i$ (the latent vector that controls background appearance) from the NeRF-based models (\nerfl{}, \nerfls{}, \nerflsd{}) since the pure gray background does not change for each instance.


\subsection{Cups}

We detail the exact cups videos that we manually selected from Objectron~\cite{objectron2020} to build our \textsc{\small cups} dataset in Table~\ref{table:cups_videos}.

\begin{table*}[t]
\centering
\begin{tabular}{c||c|c|c|c|c|c|c|c|c|c|c|c|c|c|c}
\toprule
batch & 1 & 3 & 5 & 6 & 7 & 8 & 9 & 13 & 15 & 16 & 18 & 21 & 23 & 24 & 25\\ \hline
video number & 22 & 1 & 3 & 15 & 24 & 20 & 13 & 4 & 13 & 21 & 2 & 21 & 2 & 29 & 47 \\
 &  & 30 & 14 & 16 & 28 & 36 & 23 & 9 & 15 & 28 & 4 & 28 & 38 &  &  \\
 &  &  & 39 & 25 & 40 &  & 45 & 15 & 17 & 39 & 8 & 38 &  &  &  \\
 &  &  &  & 26 & 46 &  & 48 &  & 39 & 49 & 9 &  &  &  &  \\
 &  &  &  & 34 &  &  &  &  & 48 &  & 16 &  &  &  &  \\
 &  &  &  & 42 &  &  &  &  &  &  & 18 &  &  &  &  \\
 &  &  &  &  &  &  &  &  &  &  & 25 &  &  &  &  \\
 &  &  &  &  &  &  &  &  &  &  & 30 &  &  &  &  \\
 &  &  &  &  &  &  &  &  &  &  & 31 &  &  &  &  \\
 &  &  &  &  &  &  &  &  &  &  & 33 &  &  &  &  \\
 &  &  &  &  &  &  &  &  &  &  & 35 &  &  &  &  \\
 &  &  &  &  &  &  &  &  &  &  & 47 &  &  &  &  \\
\midrule \hline
batch & 27 & 28 & 30 & 31 & 32 & 33 & 34 & 36 & 38 & 39 & 41 & 42 & 45 & 46 & 48\\ \hline
video number & 42 & 36 & 11 & 12 & 13 & 20 & 38 & 29 & 17 & 2 & 16 & 0 & 7 & 5 & 23 \\
 &  &  & 22 & 38 & 16 &  &  &  & 20 & 18 &  & 2 & 37 & 14 & 38 \\
 &  &  & 29 & 44 & 21 &  &  &  & 34 & 44 &  & 10 & 46 & 39 & 41 \\
 &  &  & 36 &  & 31 &  &  &  &  &  &  & 12 & 47 &  & 49 \\
 &  &  & 40 &  & 38 &  &  &  &  &  &  & 17 &  &  &  \\
 &  &  & 46 &  & 40 &  &  &  &  &  &  & 20 &  &  &  \\
 &  &  &  &  & 44 &  &  &  &  &  &  & 21 &  &  &  \\
 &  &  &  &  &  &  &  &  &  &  &  & 32 &  &  &  \\
\bottomrule
\end{tabular}
\caption{Videos used from Objectron.}
\label{table:cups_videos}
\end{table*}

\section{More Results}

Please see \website{} for more results. We show instance interpolation, viewpoint interpolation and extrapolation, separation, and amodal segmentation results.

{\small
\bibliographystyle{ieee_fullname}
\bibliography{objects_bib}
}

\end{document}